\documentclass{isprs} 
\usepackage{subfigure}
\usepackage{setspace}
\usepackage{amsmath}
\usepackage{geometry} 
\usepackage{epstopdf}
\usepackage[labelsep=period]{caption}  
\usepackage[british]{babel} 
\usepackage[hang]{footmisc}
\usepackage{booktabs}

\begin{document}
\nocite{*}
\title{SatOV: Restoring Spatial Priors for Training-Free Open-Vocabulary Segmentation in Remote Sensing Imagery}
\date{}


\author{
Changhao Zhao$^1$,
Haoxiang Li$^1$,
Hai Liu$^1$,
LingLin Zeng$^{1,\dagger}$
}

\address{
$^1$College of Resources and Environment, Huazhong Agricultural University, Wuhan 430070, China \\
\quad\{132, hailiu, l\_h\_x\}@webmail.hzau.edu.cn, zenglinglin@mail.hzau.edu.cn\\
}


\abstract{

Open-vocabulary semantic segmentation (OVS) of remote sensing imagery is a challenging pixel-level understanding task that demands strong generalization and adaptation to the unique spatial characteristics of remote sensing data. While existing vision-language foundation models excel in general-domain scenarios, their image-level classification design inevitably leads to the degradation of spatial priors required for high-resolution remote sensing segmentation. In particular, spatial information is degraded at two different stages of the representation pipeline: structural spatial relations are weakened during deep feature transformation, while fine-grained spatial details are lost during feature downsampling. To address these complementary deficiencies, we propose SatOV, a training-free downstream framework for open-vocabulary segmentation in remote sensing, built around a unified perspective of \textbf{spatial-prior restoration at two different stages of the representation pipeline}. Specifically, (1) \textbf{Residual QQ Attention (ResQQ)} restores structural spatial priors in deep feature representations by extracting Query-Key self-attention from an intermediate CLIP layer and fusing it with the final-layer Query-Query attention through a residual combination, thereby preserving spatially coherent relationships suppressed by the final-layer representation; and (2) \textbf{Spatially Modulated Upsampling (SatUp)} restores fine-grained spatial priors lost during downsampling by using the original high-resolution RGB image as spatial guidance and combining spatial feature modulation with guided cross-attention to reconstruct pixel-level textures and boundaries. Extensive experiments on multiple remote sensing benchmarks, including DOTA, UDD, LoveDA, and Vaihingen, demonstrate that SatOV consistently improves training-free OVS performance and achieves competitive results against existing state-of-the-art methods in both quantitative and qualitative evaluations. These results demonstrate the effectiveness of restoring spatial priors at both the representation and spatial-resolution stages for remote sensing open-vocabulary segmentation.
}


\maketitle


\section{Introduction}
 
\sloppy

Semantic segmentation of remote sensing imagery is widely demanded in agricultural monitoring, disaster assessment, and urban planning. Traditional supervised methods, however, rely on large‑scale pixel‑wise annotations and generalise poorly to unseen scenes. How to achieve high‑generalisation segmentation with low annotation dependency remains a core challenge.

Vision‑language models (VLMs) like CLIP\cite{radford2021learning} have opened new avenues for open‑vocabulary semantic segmentation (OVS). Yet CLIP is inherently designed for image‑level classification, resulting in a pronounced global semantic bias\cite{li2025segearthov} and insufficient fine‑grained spatial modelling. When applied directly to remote sensing images, a severe domain prior drift emerges – natural images typically centre on a single prominent object, whereas remote sensing scenes feature densely packed structures and anisotropic textures – fundamentally undermining segmentation performance.

Specifically, CLIP suffers from two intertwined spatial‑prior losses. First, although its intermediate Transformer layers encode rich spatial structure, the final‑layer global self‑attention progressively overwrites these priors, causing the output features to lose covariance with respect to rotation and scale. Second, the final output resolution is only 14×14 or 16×16; the abundant texture, edge, and detail priors from the original high‑resolution image are discarded during downsampling and cannot be recovered by simple interpolation.

Existing solutions fall into two main categories. One is fine‑tuning CLIP\cite{liu2024remoteclip}, which incurs prohibitive annotation costs and introduces dataset bias\cite{wysoczanska2023clipdino}. The other comprises training‑free methods\cite{karazija2025survey} that typically only replace the final‑layer attention and thus fail to handle complex spatial relationships in remote sensing. This failure stems from CLIP’s image‑level pre-training objective: its global self-attention inherently erases local spatial correlations, and tweaking only the final attention layer cannot remedy this issue. Meanwhile, alongside VLM-based models, self-supervised foundation models\cite{xiao2025foundation} have also advanced; however, they typically rely on downstream fine-tuning for deployment and thus cannot provide instant generalisation in dynamic, open-world scenarios. To recover spatial priors, we exploit two complementary sources: \textbf{(1) intermediate attention matrices}, which retain patch-wise spatial relations before they are globally collapsed, and \textbf{(2) high-resolution images}, which preserve fine-grained edges and textures that are otherwise lost under CLIP’s standard low-resolution input.

In response, this work targets a genuinely training‑free and zero‑shot segmentation paradigm – direct application to arbitrary remote sensing images without any parameter updates or fine‑tuning, and with the ability to recognise novel categories never seen during training. We propose SatOV, a training-free downstream framework for zero-shot open-vocabulary segmentation in remote sensing, which systematically addresses CLIP's spatial-prior deficiencies through through intrinsic architectural mechanisms. 

Beyond empirical gains, we analyse CLIP's limitations for remote‑sensing dense prediction through feature‑space spatial covariance. Meaningful segmentation representations require correlated features for spatially adjacent patches. Optimised for image‑level classification, CLIP's final‑layer global self‑attention gradually erases patch‑level spatial correlation. This covariance decay stems from its pre‑training objective and cannot be fully resolved by adjusting only final‑layer attention. Accordingly, we recover spatial priors from two orthogonal sources: intermediate Transformer attention matrices retaining patch relationships, and high‑resolution input imagery preserving edges and textures.

Our main contributions are summarized as follows:

\begin{itemize}
    \item We identify two complementary spatial-prior deficiencies of
    CLIP for training-free open-vocabulary segmentation in remote
    sensing imagery: the weakening of intermediate spatial relations
    and the loss of high-resolution visual details.

    \item We propose Residual Query-Query Attention (ResQQ), which
    combines final-layer query-query attention with relational attention
    extracted from an intermediate CLIP layer. This design preserves
    spatially coherent structures while retaining the semantic
    generalization of the CLIP representation, without any
    target-dataset adaptation.

    \item We propose Spatially Modulated Upsampling (SatUp), a
    remote-sensing-pretrained feature upsampler that uses the original
    high-resolution RGB image as spatial guidance. SatUp reconstructs
    dense semantic features through spatial feature modulation and
    guided cross-attention, complementing the structural prior recovered
    by ResQQ.
\end{itemize}

Note that “training‑free” in this paper strictly means \textbf{no fine‑tuning or gradient updates on downstream target datasets}; SatUp undergoes an offline general‑purpose pre‑training on Million‑AID, and its weights remain frozen during inference.

\section{Related Work}

\subsection{Vision-Language Models and Remote Sensing Foundation Models}

In recent years, vision foundation models and vision-language models have advanced rapidly. Among them, CLIP stands out by achieving coarse-grained cross-modal semantic alignment through image-text contrastive learning, demonstrating strong zero-shot capability and requiring no training or fine-tuning for downstream tasks—it can be used out of the box. In stark contrast, mainstream remote sensing foundation models predominantly adopt MIM or contrastive learning as their pre-training paradigms, with representative examples including OlmoEarth\cite{Herzog_2026_CVPR}, AlphaEarth\cite{Brown2025AlphaEarthFA}, and TESSERA\cite{feng2026tessera}. These models aim at visual representation learning rather than cross-modal semantic understanding. Although they are also pre-trained on massive unlabeled remote sensing data without human annotations, they still require fine-tuning for specific downstream tasks at deployment, falling short of the genuine zero-shot generalisation achieved by CLIP. In essence, the two paradigms differ fundamentally: one is trained on natural images and directly generalises zero-shot to remote sensing, while the other is trained on remote sensing data yet remains constrained by fine-tuning dependence, limiting its applicability in dynamic open-world scenarios.
         
\subsection{Training‑Free Open‑Vocabulary Segmentation}
The global semantic bias of CLIP severely limits its dense segmentation performance. To address this, many training‑free methods reformulate CLIP's final‑layer attention to enhance local modelling. MaskCLIP\cite{zhou2022maskclip} removes global pooling and replaces fully connected layers with convolutions to output dense feature maps. SCLIP\cite{wang2023sclip} replaces standard attention with Query‑Query attention, cutting global token interactions for stronger local focus. ClearCLIP\cite{lan2024clearclip} removes residual connections and feed‑forward networks with class‑token debiasing to strip global background. NACLIP\cite{hajimiri2025naclip} introduces neighbourhood Gaussian attention for local clustering preference. PEARL\cite{pei2026pearl} uses orthogonal transformation and smooth propagation for refinement. ProxyCLIP\cite{lan2024proxyclip} and LPOSS\cite{stojnic2025_lposs} incorporate DINO affinity matrices or Laplacian label propagation for finer local priors.

These methods are training‑free and achieve preliminary dense prediction by modifying attention or features. However, they are primarily designed for natural images with single dominant objects and clear features, overlooking the dense, anisotropic, and complex characteristics of remote sensing. They remain within CLIP's feature space, fail to address the overwriting of intermediate spatial priors, and neglect pixel‑level details from original high‑resolution images, leading to insufficient recovery for remote sensing imagery.

\subsection{Open-Vocabulary Segmentation for Remote Sensing Imagery}

Thanks to the dominance of single objects, distinct features, and moderate resolution in natural images, training‑free open‑vocabulary segmentation has made significant progress in natural scenes. However, extending these methods to dense remote sensing imagery remains underexplored, and most existing efforts still rely on training.

Cat‑Seg\cite{cho2024catseg} first adopts cost‑aggregation training to optimise CLIP features; OVRS\cite{cao2025open} further adapts this approach for remote sensing characteristics; GSNet\cite{ye2025GSNet} builds upon Cat‑Seg by incorporating DINO's fine‑grained features. These methods generally depend on training, incurring high annotation costs and introducing dataset bias that undermines CLIP's original generalisation. SegEarth‑OV\cite{li2025segearthov} addresses the resolution bottleneck by introducing FeatUp\cite{fu2024featup} as an upsampling module and removing global class‑token bias via subtraction filtering. However, FeatUp's training based on joint bilateral filtering remains neighbourhood‑weighted, failing to fully leverage the spatial structural priors embedded in the original RGB images.

Unlike the above works, we systematically analyse the two types of spatial prior loss faced by CLIP in remote sensing and propose targeted recovery solutions. Under the strict training‑free and zero‑shot setting, our method significantly outperforms existing state‑of‑the‑art approaches, offering a new and effective pathway for open‑vocabulary segmentation in remote sensing.


\section{Our Method}

\subsection{Overall Architecture}
\label{sec:overall_arch}
Existing training‑free methods primarily focus on optimising the neighbourhood attention in CLIP’s final layer, which generally works well for natural images. However, remote sensing imagery suffers from two fundamentally distinct types of spatial prior loss. We interpret these two failure modes from the perspective of feature spatial covariance.

Let $\mathbf{F} \in \mathbf{R}^{N\times D}$ denote patch features, where $N$ is the number of spatial patches and $D$ is the feature dimension. We define the spatial covariance matrix $\mathbf{\Sigma}\in\mathbf{R}^{N\times N}$:
\begin{equation}
\Sigma_{i,j} = \frac{1}{D}\sum_{d=1}^{D}\tilde{\mathbf{F}}_{i,d}\tilde{\mathbf{F}}*{j,d}
\end{equation}
where $\mathbf{\Sigma}_{i,j}$ quantifies feature correlation between patch $i$ and patch $j$. For dense segmentation, physically adjacent patches should maintain high covariance.

First, \textit{covariance erasure in final-layer attention}: CLIP intermediate layers encode strong spatial covariance among patches. Optimised for image-text alignment, the final-layer global self-attention mixes tokens across the whole image and gradually decays $\mathbf{\Sigma}$. We characterize this trend by averaging $\Sigma_{i,j}^{(l)}$ over spatially neighboring patch pairs $\mathcal{N}$ at each layer $l$. Even QQ-attention (e.g., SCLIP) only modifies the final layer and cannot retrieve covariance already discarded in forward passes. Second, \textit{irreversible high-frequency loss during down-sampling}: patch embedding compresses fine-grained edges and textures into low-dimensional tokens. Simple interpolation acts as a linear smoother and cannot restore averaged-out high-frequency signals such as building boundaries and narrow roads. These two degradation modes are theoretically orthogonal: one occurs inside the transformer feature space, the other arises from image downsampling.

\begin{figure}[htbp]
\centering
\includegraphics[width=1\linewidth]{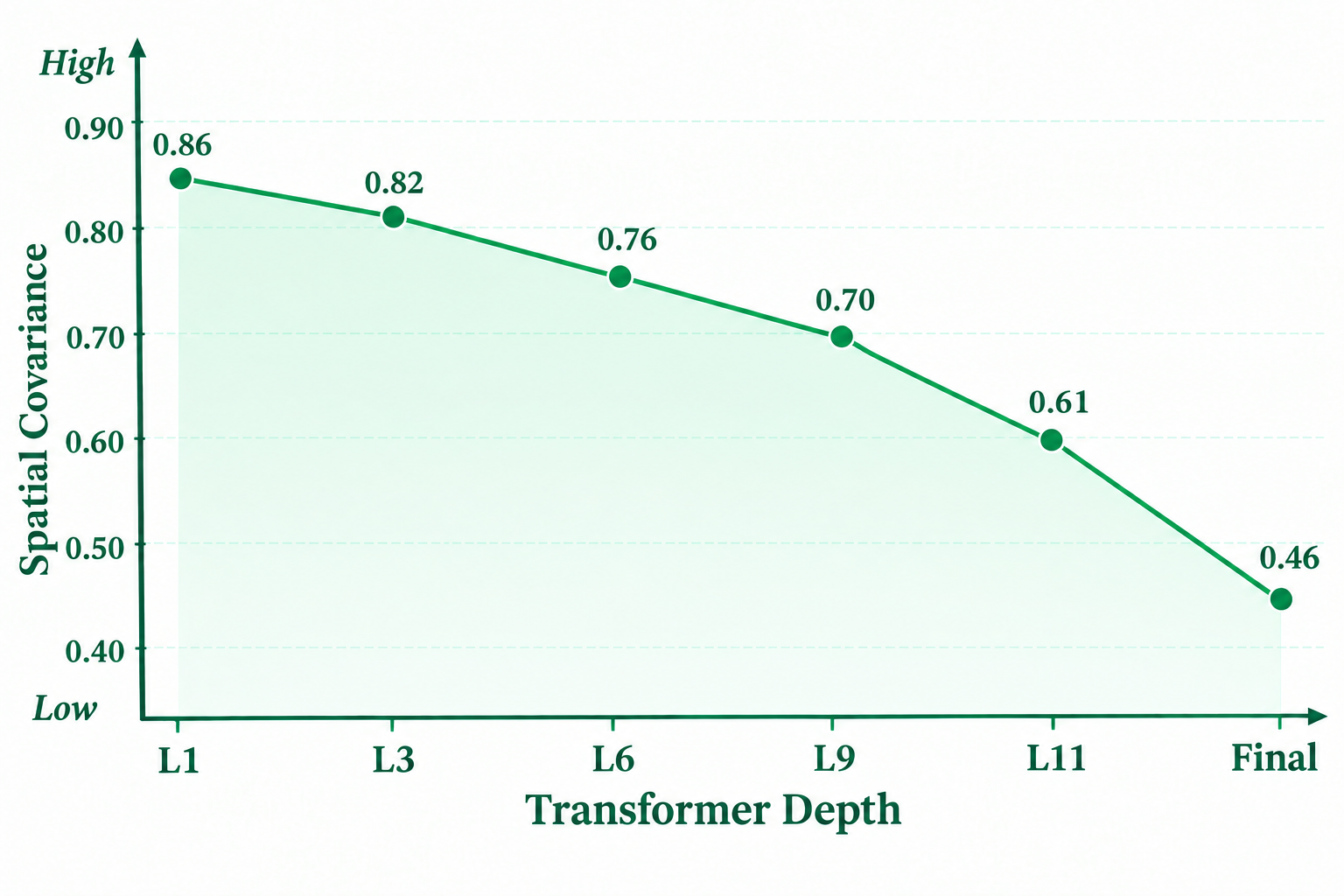}
\caption{Experimental analysis of spatial covariance across Transformer depth, showing the progressive loss of spatial detail with increasing semantic abstraction.}
\label{fig:spatial_covariance}
\end{figure}

To address these issues, ResQQ and SatUp are respectively designed to recover these two types of priors. ResQQ operates within CLIP’s feature space, extracting the overwritten structural priors from intermediate layers and injecting them into the final layer to restore spatial covariance. SatUp, on the other hand, leverages the original high‑resolution RGB image as guidance to reconstruct the pixel‑level high‑frequency details discarded during downsampling.

SatOV consists of two core components:
\textbf{Residual QQ Attention (ResQQ):} Through residual connections, it fuses the spatially covariant Query‑Key attention priors from CLIP’s intermediate layers with the Query‑Query self‑attention in the final layer, restoring the spatial structure and neighbourhood awareness of the output features.
\textbf{Spatially Modulated Upsampling (SatUp):} Guided by the original high‑resolution RGB image, it employs spatial feature modulation and cross‑attention to upsample CLIP’s coarse‑grained features to pixel‑level fine‑grained features, recovering the fine spatial details of remote sensing imagery.

\begin{figure*}[t]
    \centering
    \includegraphics[width=1\linewidth]{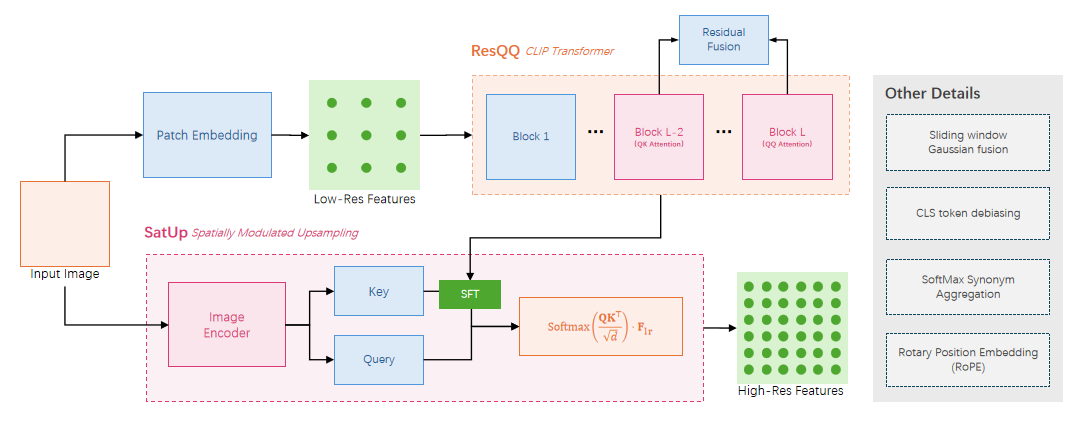}
    \caption{Left panel illustrates the restoration of structural priors within the CLIP ViT: ResQQ extracts the Query‑Key attention from the penultimate layer as the overwritten spatial prior, and fuses it with the final‑layer Query‑Query self‑attention via residual connections, recovering the spatial covariance of the low‑resolution features. Right panel demonstrates the recovery of pixel‑level detail priors: SatUp uses the original high‑resolution RGB image as a spatial index, and applies Spatial Feature Modulation (SFT) together with cross‑attention to upsample the low‑resolution semantic features back to pixel‑level space.}
    \label{fig:arch_overview}
\end{figure*}



\subsection{Dynamic Positional Encoding and Global Bias Removal}

To accommodate varying input resolutions, we dynamically interpolate the fixed positional embeddings to match the current feature map size. Since CLIP's positional embeddings are pre-trained for a specific grid size (e.g., \(14 \times 14\) or \(16 \times 16\)), directly applying them to different resolutions would cause spatial misalignment. We thus perform bicubic interpolation to resize the positional embeddings to the current feature grid dimensions:

\begin{equation}
\mathbf{P}_{\text{new}} = \mathcal{I}_{\text{bicubic}}\left(\mathbf{P}_{\text{old}},\; (H_{\text{new}}, W_{\text{new}})\right)
\label{eq:pos_embed}
\end{equation}

where \(\mathbf{P}_{\text{old}} \in \mathbf{R}^{(1+HW) \times D}\) denotes the original positional embedding matrix, \(\mathcal{I}_{\text{bicubic}}(\cdot, (H_{\text{new}}, W_{\text{new}}))\) represents the bicubic interpolation operator resizing to the target spatial dimensions, and \((H_{\text{new}}, W_{\text{new}})\) denotes the height and width of the current feature map.

In addition, the CLS token in CLIP encodes global image-level semantics learned through contrastive pre-training, which is beneficial for image classification but introduces a global bias that contaminates local patch features in dense prediction tasks. To eliminate this interference, we subtract the CLS token from every patch token:

\begin{equation}
\hat{\mathbf{F}}(\mathbf{p}) = \mathbf{F}_{\mathrm{patch}}(\mathbf{p}) - \mathbf{F}_{\mathrm{cls}},\quad \forall \mathbf{p}\in \Omega
\label{eq:debias}
\end{equation}

where \(\mathbf{F}_{\text{patch}}(\mathbf{p}) \in \mathbf{R}^D\) is the feature vector at spatial position \(\mathbf{p}\), \(\mathbf{F}_{\text{cls}} \in \mathbf{R}^D\) is the CLS token feature, and \(\Omega\) denotes the set of all patch positions. This subtraction explicitly removes the global semantic component from each patch, allowing the subsequent attention mechanisms to focus on local spatial structures rather than global context.

\subsection{Residual Query-Query Attention (ResQQ)}

The native self-attention in the final layer of CLIP is a standard global self-attention, where every token attends to all other tokens including the CLS token. This global interaction gradually overwrites the spatial structural priors encoded in the intermediate layers. Inspired by SCLIP, we replace the standard Query-Key attention with Query-Query attention in the last layer. The resulting attention is still globally computed over all patch tokens, but its affinity is determined directly by query-query similarity, reducing the semantic asymmetry introduced by the learned key projection. The standard self-attention is formulated as:

\begin{equation}
\text{Attn}(\mathbf{Q}, \mathbf{K}, \mathbf{V}) = \text{softmax}\left(\frac{\mathbf{Q}\mathbf{K}^\top}{\sqrt{d}}\right)\mathbf{V},
\end{equation}

while our adopted Query-Query self-attention in the final layer is:

\begin{equation}
\text{Attn}_{\text{self}}(\mathbf{Q}, \mathbf{Q}, \mathbf{V}) = \text{softmax}\left(\frac{\mathbf{Q}\mathbf{Q}^\top}{\sqrt{d}}\right)\mathbf{V}.
\end{equation}

Nevertheless, relying solely on the final-layer Q-Q attention discards the rich spatial structure encoded in the intermediate layers. To remedy this, inspired by ResCLIP\cite{Yang_2025_CVPR}, we propose \textbf{Residual Query-Query Attention (ResQQ)}, which injects the intermediate spatial priors into the final layer via a residual connection.

Specifically, during the forward pass through the the penultimate layer (the $(L-1)$-th layer in 1-index, i.e., the $(L-2)$-th layer in 0-index), we extract its Query-Key self-attention matrix. This layer has not yet been contaminated by the global self-attention of the final layer, thus its attention distribution preserves meaningful spatial relationships among patches. Following our implementation, we compute the Query \(\mathbf{Q}_{L-1}\) and Key \(\mathbf{K}_{L-1}\) from this layer and obtain the cross-attention:

\begin{equation}
\mathbf{A}_{\text{mid}} = \text{softmax}\left(\frac{\mathbf{Q}_{L-1} \mathbf{K}_{L-1}^\top}{\sqrt{d_{\text{mid}}}}\right),
\end{equation}

where \(d_{\text{mid}}\) is the head dimension of that layer. Simultaneously, we compute the final-layer Query-Query attention matrix \(\mathbf{A}_{\text{last}}\). We then fuse the two attention matrices via a residual combination:

\begin{equation}
\mathbf{A}_{\mathrm{fused}} = (1-\alpha)\cdot \mathbf{A}_{\mathrm{last}} + \alpha \cdot \mathbf{A}_{\mathrm{mid}}
\end{equation}

where \(\alpha\) is the residual fusion weight (set to 0.5 by default in our code). When the number of attention heads differs between the two layers, we align \(\mathbf{A}_{\text{mid}}\) by averaging over the head dimension to obtain a single aggregated attention map. This aggregated matrix is then replicated along the head dimension to match the head count of the final layer. The fused matrix is renormalized and used to weight the final‑layer Value vectors, producing the output of the attention module.

With this design, ResQQ combines the query-driven affinity of Q-Q attention with the spatial structural relations preserved in the intermediate layer. The residual fusion therefore improves the spatial coherence of the output features while retaining the semantic representation of CLIP.

\subsection{Spatially Modulated Upsampling (SatUp)}

After the ResQQ attention reconstruction, the CLIP output features regain spatial structural priors. However, the spatial resolution remains only \(14\times14\) or \(16\times16\). The rich texture, edge, and detail priors from the original high-resolution remote sensing images are discarded during CLIP's downsampling and cannot be recovered within its feature space. To address this, we design \textbf{SatUp}, a plug-and-play upsampling module tailored for remote sensing, which leverages the original high-resolution RGB image as a guide to restore low-resolution semantic features to pixel-level space.

\begin{figure*}
    \centering
    \includegraphics[width=1\linewidth]{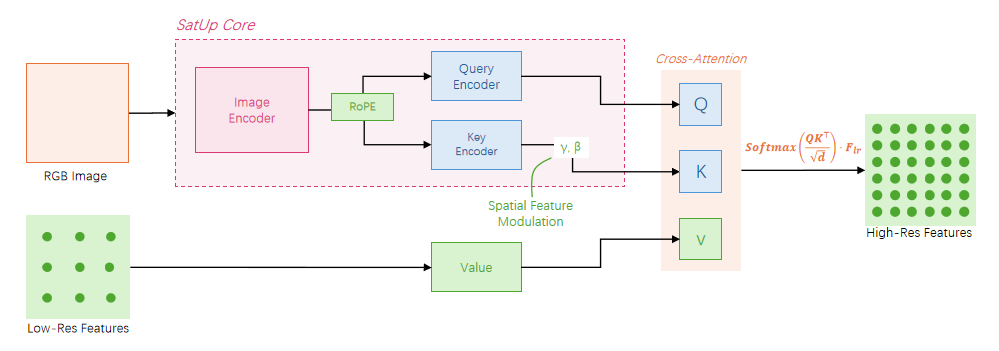}
    \caption{SatUp takes low-resolution CLIP features and the original high-resolution RGB image as inputs. It encodes the image into spatial queries, applies rotary positional encoding, and performs cross-attention with SFT-modulated keys and semantic values, producing pixel-aligned high-resolution features.}
    \label{fig:placeholder}
\end{figure*}

Unlike Neighborhood Attention-based upsampling approaches such as FeatUp, NAF\cite{chambon2025nafzeroshotfeatureupsampling}, and AnyUp\cite{wimmer2026anyup}, SatUp employs \textbf{spatial feature modulation} and \textbf{global cross-attention} as core mechanisms to extract spatial priors from the high-resolution image and drive fine-grained pixel-level reconstruction. It is worth emphasising that our definition of "training-free" refers to the inference stage, where no fine-tuning or gradient updates are performed on any downstream dataset. The offline pre-training of SatUp is consistent with the pre-training nature of the CLIP visual encoder—both build general-purpose, category-agnostic capabilities with frozen weights during inference, without task-specific adaptation.

\textbf{Rotary Position Embedding (RoPE).} Unlike directly interpolating CLIP's fixed positional embeddings, SatUp incorporates a Rotary Position Embedding specifically designed for the overhead-view characteristics of remote sensing. For each spatial position \((i,j)\) on the feature map, we first generate normalised coordinates:

\begin{equation}
\mathbf{c}_{i,j} = \left( \frac{i}{H}, \frac{j}{W} \right) \in [0,1]^2
\end{equation}

The coordinate matrix is then multiplied with a learnable frequency matrix \(\boldsymbol{\theta} \in \mathbf{R}^{2 \times d}\) to obtain rotation angles, and a rotation is applied to each token feature \(\mathbf{x} \in \mathbf{R}^d\):

\begin{equation}
\text{RoPE}(\mathbf{x}, \mathbf{c}) = \mathbf{x} \odot \cos(\mathbf{c} \cdot \boldsymbol{\theta}) + \text{rotate\_half}(\mathbf{x}) \odot \sin(\mathbf{c} \cdot \boldsymbol{\theta})
\end{equation}

where \(\text{rotate\_half}(\mathbf{x}) = [-x_{d/2+1}, \dots, -x_d, x_1, \dots, x_{d/2}]\). This encoding explicitly injects spatial position information, enabling the subsequent attention to perceive rotation and scale variations.

\textbf{Guided Cross-Attention.} This is the core forward design of SatUp for upsampling. We encode the high-resolution RGB image into spatial features \(\mathbf{F}_{\text{img}} \in \mathbf{R}^{B \times d \times H \times W}\), and generate Query and Key from it:

\begin{equation}
\mathbf{Q} = \Phi_q(\mathbf{F}_{\mathrm{hr}}),\qquad \mathbf{K} = \Phi_k(\mathbf{F}_{\mathrm{hr}})
\end{equation}

Meanwhile, the low-resolution semantic features \(\mathbf{F}_{\text{clip}} \in \mathbf{R}^{B \times d \times h \times w}\) from CLIP are mapped to Value via an encoder:

\begin{equation}
\mathbf{V} = \Phi_v(\mathbf{F}_{\mathrm{lr}})
\end{equation}

The cross-attention is then computed as:

\begin{equation}
\mathbf{F}_{\mathrm{up}} = \mathrm{Softmax}\left(\frac{\mathbf{Q}\mathbf{K}^{\top}}{\sqrt{d}}\right)\mathbf{V}
\end{equation}

In practice, $\mathbf{V}$ is first linearly projected to the same dimension as $\mathbf{Q}$ and $\mathbf{K}$ for attention computation, and the output is then projected back to the original semantic dimension.

The physical intuition is that the rich edge and texture cues in the high-resolution image serve as a "spatial index" to precisely query and retrieve the corresponding low-resolution semantic features, enabling pixel-level detail reconstruction. This process involves only linear projections and matrix multiplications, without any iterative optimisation.

\textbf{Spatial Feature Modulation (SFT).} To further enhance the guidance of high-resolution images over semantic features, we introduce Spatial Feature Modulation in the Key branch. Inspired by global modulation methods such as AdaIN\cite{huang2017adain} and FiLM\cite{perez2018film}, but overcoming their lack of spatial sensitivity, SFT\cite{wang2018sftgan} generates independent modulation parameters for each spatial position from the low-resolution semantic features:

\begin{equation}
\gamma,\beta = \mathcal{G}(\mathbf{F}_{\mathrm{lr}})
\end{equation}

where \(\mathcal{G}\) is implemented by two convolutional layers. The guidance features are then modulated as:

\begin{equation}
\mathbf{F}_{\mathrm{mod}} = \gamma \odot \mathrm{GroupNorm}(\mathbf{F}_{\mathrm{hr}}) + \beta
\end{equation}

Notably, the GroupNorm here contains no learnable affine parameters (\texttt{affine=False}), ensuring that the modulation is entirely driven by the current input image rather than relying on pre-trained statistics.

\textbf{Training and Loss Function.} SatUp adopts a zero-shot self-supervised training strategy without any semantic annotations. Following the training paradigms of JAFAR and LoftUp\cite{Huang_2025_ICCV}, we leverage feature consistency and continuous reconstruction between image pairs of different resolutions as supervision signals. During training, two images of the same scene at resolutions \(224 \times 224\) and \(112 \times 112\) are fed as input. SatUp is required to upsample the low-resolution features to high-resolution space and maintain semantic consistency with the ground-truth high-resolution features. The supervision loss combines cosine similarity and L2 loss:
\begin{equation}
\mathcal{L} = \frac{1}{N}\sum_{i=1}^{N}\left[\lambda_1\left(1 - \cos(\mathbf{f}_{i}^{\mathrm{up}},\mathbf{f}_{i}^{\mathrm{hr}})\right) + \lambda_2\|\mathbf{f}_{i}^{\mathrm{up}} - \mathbf{f}_{i}^{\mathrm{hr}}\|_{2}\right]
\end{equation}

The training data is sampled from a subset of Million-AID. Using a single RTX 4080 GPU with a batch size of 5, we train for 25,000 steps, achieving a final convergence loss of approximately 0.3. After training, the SatUp weights are completely frozen during inference, without any fine-tuning on downstream datasets, preserving the training-free nature of the entire framework.

\subsection{Sliding-Window Inference with Gaussian Fusion}

Since CLIP is pre-trained with a fixed resolution of $224\times224$, directly resizing high-resolution remote sensing images would result in significant loss of fine-grained details. To address this, we adopt a sliding-window strategy that partitions the input image into overlapping $224\times224$ patches, each fed independently into SatOV for inference. To mitigate stitching artefacts along window boundaries, we apply a 2D Gaussian weight mask that assigns higher confidence to the central region of each window while smoothly decaying towards the edges:

\begin{equation}
\mathcal{M}(u,v) = \exp\left(-\frac{(u-\mu_u)^2 + (v-\mu_v)^2}{2\sigma^2}\right)
\end{equation}

The predictions from all windows are then weighted by this mask, accumulated, and normalised to produce a globally consistent segmentation map with smooth transitions.

\subsection{SoftMax Synonym Aggregation}
SatOV matches pixel-level visual features with frozen CLIP text embeddings.
Since a single prompt often fails to describe a category comprehensively, we
adopt a synonym ensemble strategy: each category $c_k$ is associated with a
set of synonyms $\mathcal{S}_k=\{s_k^1,\dots,s_k^{M_k}\}$, and for each
synonym $s_k^m$ we apply a template set $\mathcal{T}$ and encode the
prompted sentences with a frozen CLIP text encoder. The resulting embeddings
are averaged and $\ell_2$-normalized to obtain $\tilde{\mathbf{e}}_k^m$.
Rather than averaging these embeddings, which dilutes strong evidence, we
reweight them by their mutual agreement,
\begin{equation}
    w_k^m
    = \frac{\exp\!\left(\tau_s \sum_{n}
      \cos(\tilde{\mathbf{e}}_k^m,\tilde{\mathbf{e}}_k^n)\right)}
      {\sum_{j}\exp\!\left(\tau_s \sum_{n}
      \cos(\tilde{\mathbf{e}}_k^j,\tilde{\mathbf{e}}_k^n)\right)},
\end{equation}
and form the category prototype
$\mathbf{e}_k=\sum_{m} w_k^m\,\tilde{\mathbf{e}}_k^m$, $\ell_2$-normalized.
As $\tau_s$ grows, the weights concentrate on the synonym most consistent
with the ensemble; as $\tau_s$ shrinks, they approach a uniform mean. The
prototypes are stacked into
$\mathbf{W}=[\mathbf{e}_1,\dots,\mathbf{e}_K]$ and matched with pixel
features scaled by a learnable temperature $\tau$ to produce the
segmentation logits. This synonym ensemble requires no fine-tuning and
consistently improves the robustness of zero-shot classification across
remote sensing categories.

\section{Experiments}

To comprehensively evaluate the effectiveness of SatOV, we design three sets of experiments. First, we compare the complete SatOV framework against state-of-the-art training-free methods in an end-to-end manner, examining its overall advantage on remote sensing segmentation tasks. Second, we fix the CLIP visual encoder and replace only the upsampler module to assess the standalone performance gain brought by SatUp. Finally, we conduct systematic ablation studies to isolate and quantify the individual contributions of each key component, including ResQQ and SatUp, to the final performance.

\subsection{Datasets}

\subsubsection{Pre-training dataset for SatUp.} To support offline pre-training of the upsampling module, we use the Million-AID\cite{Long2021DiRS} remote sensing dataset, which comprises 1,000,000 images covering 25 typical land-cover categories.

\subsubsection{Downstream evaluation datasets.} To comprehensively evaluate the zero-shot generalisation capability of SatOV under the training-free setting, we conduct experiments on four publicly available remote sensing segmentation benchmarks. These datasets cover diverse imaging platforms including unmanned aerial vehicles (UAV), high-resolution satellites, and multispectral satellites, and involve various task scenarios such as urban land-cover mapping, dense object segmentation, and agricultural field extraction. Table~1 summarises the characteristics of each dataset.
\begin{table}[htbp]
\centering
\label{tab:datasets}
\begin{tabular}{l c c}
\toprule
\textbf{Dataset} & \textbf{Imaging Platform} & \textbf{Primary Type} \\
\midrule
UDD & UAV & Urban \\
DOTA & Satellite & Dense objects \\
LoveDA & Satellite & Farmland and Urban \\
Vaihingen & Aerial (no blue band) & Building and House \\
\bottomrule
\end{tabular}
\caption{Overview of evaluation datasets.}
\end{table}
\begin{table*}[!t]
\centering
\begin{tabular}{l c c c c c}
\toprule
\textbf{Model} & \textbf{UDD} & \textbf{DOTA} & \textbf{LoveDA} & \textbf{Vaihingen} & \textbf{Avg.} \\
\midrule
\multicolumn{6}{c}{\textit{Training-Free Methods}} \\
MaskCLIP \textsubscript{ECCV baseline} & 32.4 & 25.1 & 27.8 & 23.4 & 27.2 \\
SCLIP \textsubscript{ECCV} & 38.7 & 29.3 & 30.4 & 29.1 & 31.9 \\
ClearCLIP \textsubscript{ECCV} & 41.2 & 31.0 & 31.6 & 35.5 & 34.8 \\
LPOSS \textsubscript{CVPR} & 41.8 & 33.9 & 32.4 & 38.0 & 36.5 \\
SegEarth-OV \textsubscript{CVPR} & 45.3 & 40.3 & 36.9 & 42.3 & 41.2 \\
\textbf{SatOV} \textsubscript{Ours} & \textbf{45.9} & \textbf{40.4} & \textbf{39.8} & \textbf{47.3} & \textbf{43.3} \\
\midrule
\multicolumn{6}{c}{\textit{Additional Training-based Methods} (w/o target dataset training)} \\
Cat-Seg \textsubscript{CVPR} & 31.9 & 27.8 & 20.1 & 24.5 & 29.4 \\
OVRS \textsubscript{TGRS} & 37.2 & 31.3 & 28.7 & 29.2 & 31.6 \\
GSNet \textsubscript{AAAI} & 38.2 & 29.5 & 27.0 & 30.6 & 31.3 \\
\bottomrule
\end{tabular}
\caption{Performance comparison of different methods on four remote sensing datasets. The primary comparison is conducted among training-free methods under the same zero-shot evaluation setting.}
\end{table*}

\begin{figure*}[!t]
    \centering
    \includegraphics[width=1.0\linewidth]{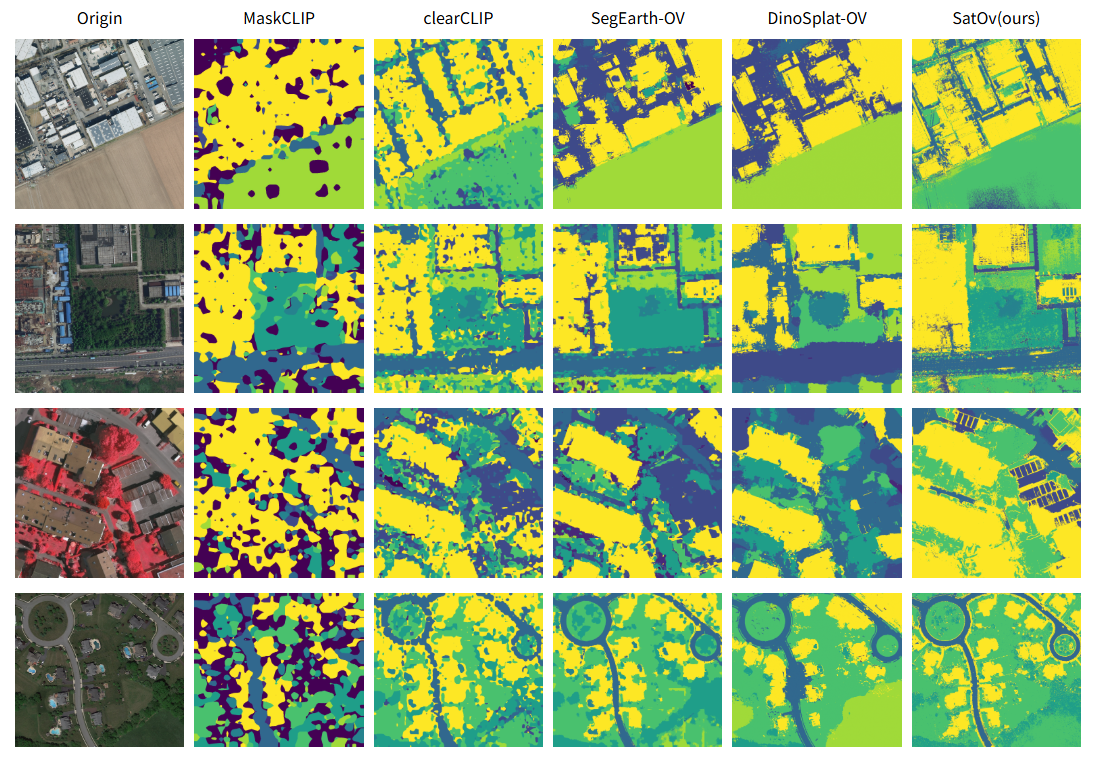}
    \caption{For the segmentation mask comparison on the DOTA dataset with densely packed objects, it can be observed that MaskCLIP predictions are largely semantically correct but completely lack spatial structure. ClearCLIP shows some improvement yet still suffers from internal holes. In contrast, SatOV accurately delineates the boundaries of densely arranged buildings and narrow roads, demonstrating the effectiveness of ResQQ and SatUp in jointly recovering spatial priors.}
\end{figure*}
\begin{figure*}[t]
    \centering
    \includegraphics[width=1.0\linewidth]{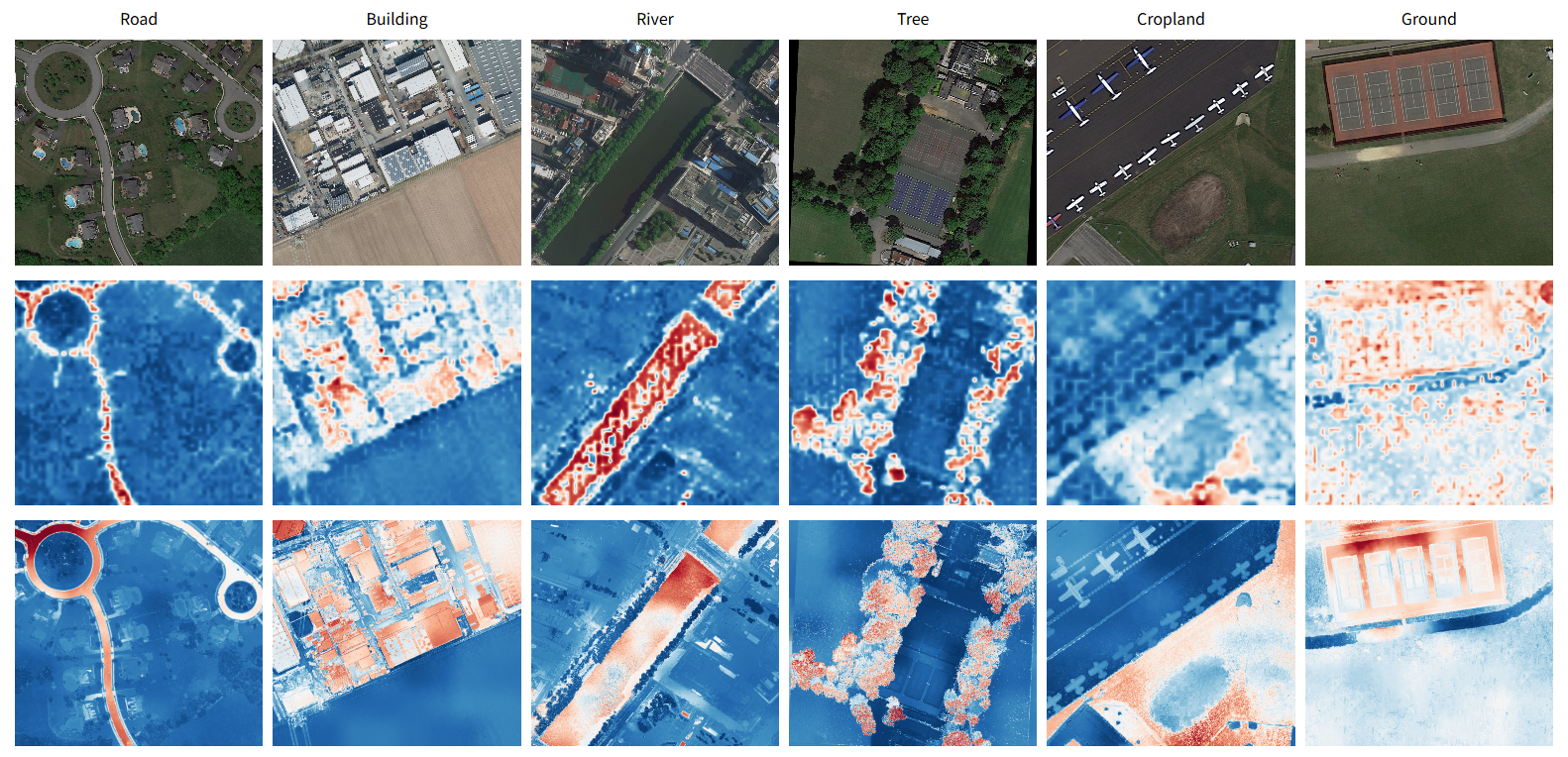}
    \caption{These are binary classification attention maps. The second row shows ClearCLIP outputs while the third row shows our SatOV. Significant differences arise from the resolution bottleneck, and SatOV achieves true pixel-level fine-grained prediction.}
\end{figure*}

For all datasets, We mainly select dense‑image subsets with high resolution (greater than 1000×1000 pixels) to better challenge the segmentation capability. Notably, the Vaihingen dataset lacks the blue spectral band, so we adapt the input channels during preprocessing by using all available bands. Since SatUp is pre‑trained on three‑channel RGB imagery, we perform channel‑wise averaging projection to map multi‑band Vaihingen inputs to 3‑channel feature representations for module compatibility. All models are evaluated directly in a zero-shot manner without any fine-tuning or parameter updates on the target datasets, ensuring a fair assessment of their training-free generalisation performance.

For the DOTA dataset, we adopt the processing pipeline from SAMRS\cite{SAMRS} to convert this object‑detection dataset into a semantic‑segmentation dataset. Among Vaihingen and Potsdam, we choose the Vaihingen dataset due to its higher spatial resolution.

\subsection{Comparative Experiments of SatOV}

Since our primary goal is training-free segmentation, we mainly compare against training-free methods. We adopt MaskCLIP as the baseline and include ClearCLIP, LPOSS, SegEarth-OV (with FeatUp) and DinoSplat-OV(with Dino\cite{Caron_2021_ICCV} and GSUP) for comparison. Although not our main focus, we also report results of training-based methods including Cat-Seg, OVRS, and GSNet, all of which are only pre-trained on RS sub-dataset of ImageNet, to examine their generalisation under lightweight training.

Experiments are conducted on four remote sensing datasets: UDD\cite{zhang2018udd}, DOTA\cite{xia2018dota}, LoveDA\cite{wang2021loveda}, and Vaihingen\cite{cramer2010vaihingen}. We use mean Intersection over Union (mIoU) as the evaluation metric. Table~2 presents the quantitative results.

As shown in Table~2, SatOV achieves the best performance across all datasets, outperforming the strongest competitor SegEarth-OV by nearly 2 percentage points on average. ClearCLIP, which only modifies the final layer without fusing intermediate features, produces noticeable semantic holes in the predictions. LPOSS introduces DINO affinity and Laplacian propagation, achieving moderate improvements; however, DINO is trained on natural images, and its features are far less effective than the pixel-wise fine-grained features recovered by SatUp. SegEarth-OV adopts FeatUp for upsampling, but its training remains neighbourhood-weighted and neglects the spatial covariance in remote sensing, leading to degraded performance in dense scenes. The effectiveness of SatOV demonstrates that in training-free open-vocabulary segmentation, explicitly recovering CLIP's missing spatial priors from both structural (via ResQQ) and detail (via SatUp) perspectives is more fundamental and efficient than merely modifying attention or introducing external affinity matrices such as DINO (as in LPOSS). 
As for Cat-Seg, OVRS, and GSNet, although they are trained on remote sensing data, their reliance on supervision limits generalisation and yields inferior zero-shot performance compared to our training-free framework. The results that their reported scores are not directly comparable due to different training protocols.

In addition to multi-class semantic segmentation tasks, we also conducted visualization for single-class extraction tasks, covering \textbf{Road, River, Building, vegetation, Cropland, and Ground}. The visualization clearly demonstrates that SatUp achieves fine-grained boundary extraction for categories such as roads, rivers, and buildings. In contrast, ClearCLIP barely has the capability of fine boundary extraction, which is the most essential requirement in remote sensing.

\subsection{Comparative Evaluation of SatUp}

\begin{figure}[!htbp]
    \centering
    \includegraphics[width=1\linewidth]{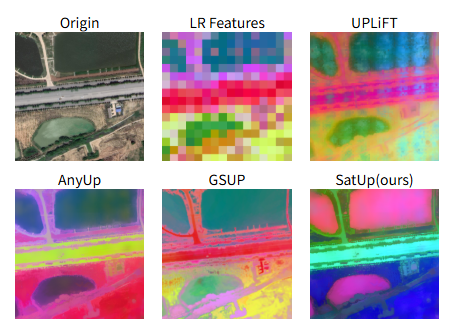}
    \caption{Visualization of feature reconstruction across different upsampling methods. It can be observed that SatUp achieves the best fine-grained reconstruction.}
    \label{fig:placeholder}
\end{figure}
\begin{table}[!htbp]
\centering
\label{tab:satup_comparison}
\begin{tabular}{l c c}
\toprule
\textbf{Model} & \textbf{Pre-training Data} & \textbf{UDD}  \\
\midrule
+DINO \textsubscript{CVPR} & ImageNet & 41.8  \\
+FeatUp \textsubscript{ICLR} & ImageNet & 45.3  \\
+AnyUp \textsubscript{ICLR} & ImageNet & 44.2  \\
\textbf{+SatUp} \textsubscript{Ours} & Million-AID & \textbf{45.9}  \\
\bottomrule
\end{tabular}
\caption{Performance comparison of different upsamplers on the UDD dataset. Taking DINO as the baseline, all methods yield performance improvements, yet SatUp achieves the best results for remote sensing tasks.}
\end{table}
In this section, we isolate the contribution of the upsampling module by fixing the CLIP features enhanced by ResQQ as the shared input representation, and evaluate the performance of different upsamplers. Notably, the compared methods—AnyUp and FeatUp—are pre-trained on natural images (ImageNet), while SatUp is specifically designed for remote sensing and pre-trained on Million-AID.

As shown in the Table~3, SatUp achieves the best overall performance among the compared upsampling methods. FeatUp relies on JBU-based training, while AnyUp employs neighbourhood attention modules such as NATTEN. Although these approaches yield improved local sensitivity, they essentially perform interpolation based on feature similarity and fail to effectively exploit the spatial priors embedded in the original RGB images. Moreover, FeatUp requires compilation of CUDA operators, which severely limits its deployability on Windows and Mac platforms. In contrast, SatUp relies solely on standard cross-attention, offering superior portability and practical applicability across different environments.

\subsection{Ablation Studies}

To systematically quantify the contribution of each component in SatOV, we conduct two sets of ablation experiments. The first set examines the impact of individual modifications within the CLIP reconstruction pipeline, while the second evaluates the plug-and-play effectiveness of SatUp under a fixed ResQQ backbone.

\subsubsection{CLIP-side Reconstruction Ablation}

We progressively remove or disable each component in the CLIP feature reconstruction pipeline to assess its individual contribution. The results are summarised in Table~4.

\begin{table}[htbp]
\centering

\label{tab:ablation_full}
\begin{tabular}{lcc}
\toprule
\textbf{Configuration} & \textbf{DOTA} & \textbf{$\nabla$} \\
\midrule
Full SatOV                        & 40.4 & --- \\
\quad $-$ SatUp                   & 34.8 & $\nabla 5.6$ \\
\quad $-$ ResQQ                   & 32.8 & $\nabla 2.0$ \\
\quad $-$ CLS token debiasing     & 30.9 & $\nabla 1.9$ \\
\quad $-$ FFN and Residual Removing     & 30.0 & $\nabla 0.9$ \\
\quad $-$ Query--Query attention  & 29.0 & $\nabla 1.0$ \\
\quad $-$ Sliding window \& Gaussian fusion & 28.4 & $\nabla 0.6$ \\
\bottomrule

\end{tabular}
\caption{Ablation study on the proposed components. Starting from the full SatOV model, we progressively remove each component to quantify its individual contribution. All models are evaluated under the same training-free, zero-shot setting on the DOTA dataset.}
\end{table}

As shown in Table~4, each component contributes positively to the final performance. Notably, CLS token debiasing and Q-Q attention bring the most substantial gains, highlighting that mitigating CLIP's inherent global semantic bias is a critical prerequisite for dense segmentation. Removing the FFN also yields a non-trivial improvement, consistent with ClearCLIP's observation that FFN layers tend to propagate global information across patches. The combination of all components achieves the best result, validating their complementary nature. When further combined with SatUp, the performance leaps to 40.4 mIoU, demonstrating that structural recovery alone is insufficient without pixel-level detail refinement.

\subsubsection{SatUp-side Plug-and-Play Ablation}

\begin{figure}[!htbp]
    \centering
    \includegraphics[width=1\linewidth]{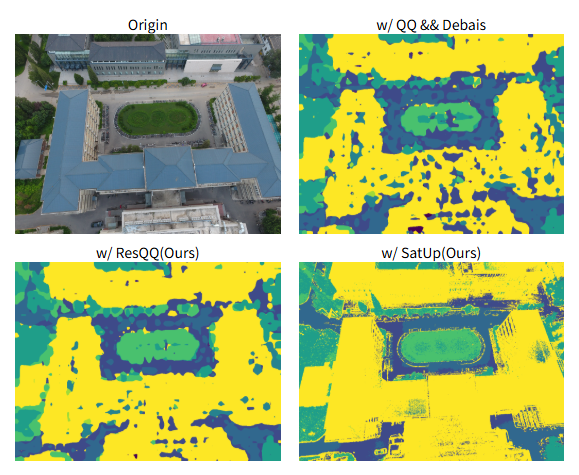}
    \caption{Visualization results for different ablation configurations. It can be seen that ResQQ can recover CLIP features to a certain extent, while SatUp leverages guidance from the original image to achieve full pixel-level granularity.}
    \label{fig:placeholder}
\end{figure}

To evaluate the plug-and-play effectiveness of SatUp, we fix the complete ResQQ-enhanced CLIP backbone and replace SatUp with alternative upsamplers or remove it entirely. The results are presented in Table~3.

Removing the upsampling module leads to a dramatic performance drop of 5.2 percentage points (from 40.4 to 34.8), indicating that simply increasing resolution via bilinear interpolation is detrimental to semantic segmentation. Among the compared methods, SatUp consistently outperforms AnyUp and GSUP\cite{zhao2026dinosplatov}, demonstrating its superiority in leveraging high-resolution RGB guidance to recover fine-grained details. GSUP, based on Gaussian splatting, achieves moderate improvement but still lags behind SatUp, likely due to its limited capability in capturing complex spatial structures. AnyUp, which relies on neighbourhood attention, shows competitive performance but tends to generate over-smoothed results and erase small ground objects. These results confirm that SatUp serves as an effective and portable plug-in module, and its design philosophy of decoupling spatial prior recovery from the visual encoder offers clear advantages over alternative upsampling strategies.
\section{Conclusion and Discussion}

\subsection{Discussion}

This work demonstrates that spatial priors for open-vocabulary segmentation can be effectively acquired through general-purpose pre-training on million-scale remote sensing datasets, without requiring any fine-tuning or gradient updates for downstream tasks. Moreover, as a plug-and-play module, SatUp is designed to be decoupled from the upstream visual encoder, facilitating the integration of spatial priors across different feature extractors.

\textbf{Potential extension beyond CLIP.} The proposed spatial prior module, SatUp, is not inherently tied to CLIP's specific architecture and can potentially be transferred to other visual encoders such as DINO and MIM-based models. This suggests that spatial-prior restoration could be incorporated into various training-free open-vocabulary segmentation paradigms, including the ``visual foundation model + text alignment'' paradigm, affinity propagation, and conditional random fields (CRF). Our concurrent work, DinoSplat-OV\cite{zhao2026dinosplatov}, which builds upon DINOv3 rather than CLIP, provides a practical example of exploring this broader direction.

\textbf{Temporal and Spectral priors.} Current mainstream methods predominantly rely on RGB channels, yet remote sensing imagery uniquely offers rich multi-spectral information (e.g., NIR, SWIR) and multi-temporal sequences, which encode abundant spatial and semantic cues. Exploring how to effectively distil these spectral and temporal priors into vision-language models remains a promising research direction.

\textbf{Bag-of-words spatial optimisation.} The performance upper bound of current open-vocabulary segmentation is constrained by the quality of the bag-of-words embedding space of CLIP or DINOv2. There is still no consensus on which vocabulary yields superior segmentation results or how to optimally combine different textual descriptions, leaving substantial room for further investigation. Although we employ synonym aggregation to alleviate this issue, better methods are worth investigating.

\subsection{Conclusion}

In this work, we proposed SatOV, a training-free framework for open-vocabulary semantic segmentation in remote sensing imagery. By identifying the degradation of spatial priors in CLIP-based representations, we introduced two complementary components, ResQQ and SatUp, to restore spatial information at different stages. ResQQ restores structural spatial relations by integrating intermediate Query-Key self-attention with final-layer Query-Query attention, while SatUp reconstructs fine-grained spatial details through spatially guided feature upsampling. Extensive experiments on DOTA, UDD, LoveDA, and Vaihingen demonstrate the effectiveness of the proposed spatial-prior restoration strategy across multiple remote sensing benchmarks.

ResQQ and SatUp complement each other by restoring spatial priors at two distinct stages of the representation pipeline: internal structural relations within the visual representation and fine-grained details at the spatial-resolution stage.

\section{Acknowledgments}
 This work was supported  by the National Natural Science Foundation of China (No.42571434).

\bibliography{ISPRSguidelines_authors}

\end{document}